\documentclass[]{style/ceurart}
\usepackage{listings}
\usepackage{longtable}
\usepackage{placeins}
\begin{document}

\copyrightyear{2026}
\copyrightclause{Copyright for this paper by its authors.
  Use permitted under Creative Commons License Attribution 4.0
  International (CC BY 4.0).}

\conference{CLEF 2026 Working Notes, 21 -- 24 September 2026, Jena, Germany}

\title{DS@GT ARC at AnimalCLEF 2026: Species-Aware Graph Construction for Multi-Species Animal Re-Identification}

\author[1]{Evan Sinclair Smith}[
    orcid=0009-0007-9097-275X,
    email=esmith446@gatech.edu,
]
\cormark[1]

\author[1]{Anthony Miyaguchi}[
orcid=0000-0002-9165-8718,
email=acmiyaguchi@gatech.edu,
]

\author[1]{Snigdha Palamari}[
    orcid=0009-0001-1729-0396,
    email=spalamari3@gatech.edu,
]

\author[1]{Danté Evangelista}[
    orcid=0009-0005-7587-6271,
    email=devangelista8@gatech.edu,
]

\address[1]{Georgia Institute of Technology, North Ave NW, Atlanta, GA 30332}
\cortext[1]{Corresponding author.}

\begin{abstract}
Automated individual animal re-identification is essential for large-scale biodiversity monitoring; however, field imagery complicates separating identity cues from nuisance variation in pose, illumination, background, resolution, and species-specific morphology. The DS@GT ARC submission to AnimalCLEF 2026 introduces a multi-species image-clustering system for re-identifying Eurasian lynx, fire salamanders, loggerhead sea turtles, and Texas horned lizards. Instead of relying on a single descriptor or nearest-neighbor retrieval, this approach formulates re-identification as species-aware graph construction over candidate image pairs. The pipeline integrates tailored preprocessing, global candidate retrieval, LightGlue-based local verification with multiple keypoint families, LightGBM pair scoring, conservative edge admission, and Leiden community detection. This design directly addresses a primary failure mode of clustering-based re-identification: high-scoring false pairs that act as bridge edges and merge distinct individuals through transitive closure. Across species, ablation studies demonstrate that local feature support, foreground-aware preprocessing, and species-specific backbone selection enhance pair evidence, while graph operating points determine the trade-off between fragmentation and over-merging. The selected submission achieved a public ARI of 0.733 and a private ARI of 0.674, ranking fifth among 230 teams. These results indicate that robust wildlife re-identification requires not only strong visual representations but also calibrated integration of global similarity, local identity markings, neighborhood context, and graph-level constraints. The code can be found at \url{https://github.com/dsgt-arc/animalclef-2026}.
\end{abstract}


\maketitle
\thispagestyle{empty}
\pagestyle{empty}

\section{Introduction}

Individual animal identification is a central task in biodiversity monitoring because repeated sightings of the same animal support estimates of population size, movement, and behavior. In this study, we developed a multi-stage computer vision system for image-based individual re-identification in AnimalCLEF 2026, a field-style clustering challenge hosted through Kaggle and described in the official AnimalCLEF overview \cite{animal-clef-2026,animalclef2026overview}. 
The challenge provides images from four species with scientific names: Lynx lynx (Eurasian lynx) \cite{picek2026czechlynx}, Salamandra salamandra (fire salamander), Caretta caretta (loggerhead sea turtle) \cite{adam2024seaturtleid2022}, and Phrynosoma cornutum (Texas horned lizard) \cite{biffi2025texas}. 
The competition evaluated submissions with the Adjusted Rand Index (ARI), which measures agreement between predicted clusters and ground-truth identities while penalizing both over-splitting one individual into multiple clusters and over-merging different individuals into the same cluster \cite{animal-clef-2026,hubert1985comparing}.

\begin{figure}
    \centering
    \includegraphics[width=1\linewidth]{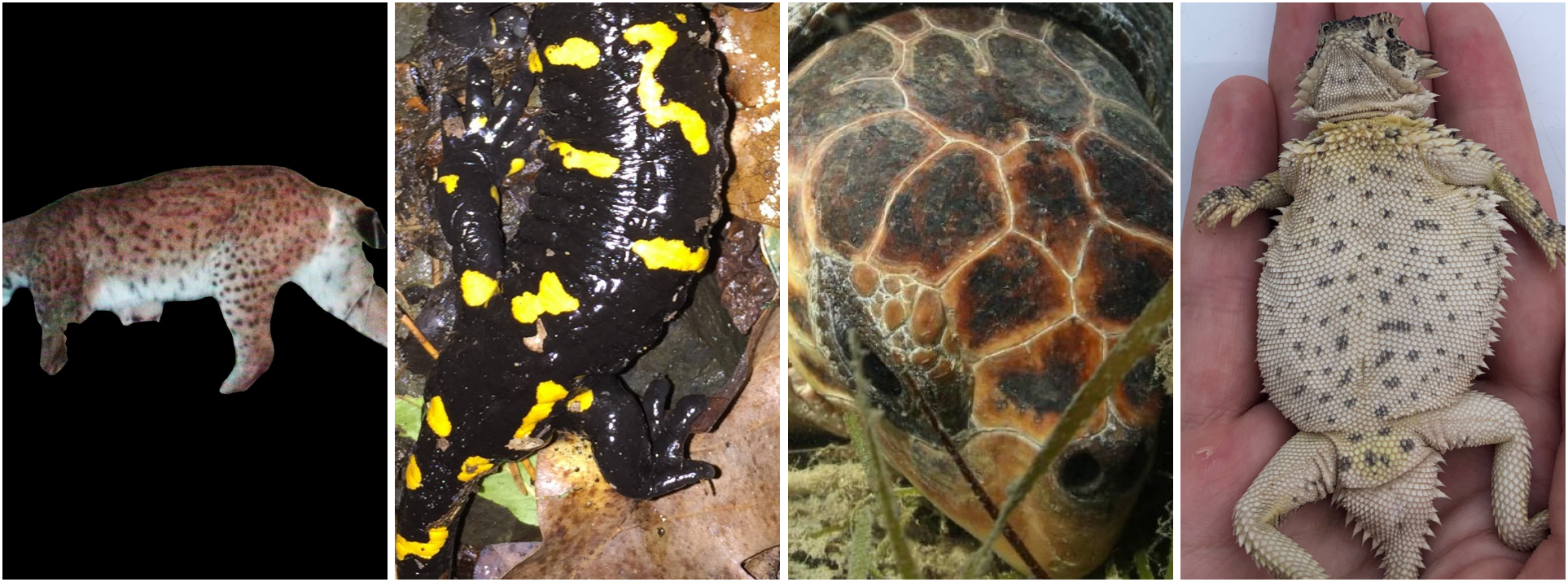}
    \caption{Sample images from each of the animals in the dataset: Lynx, Salamander, Turtle, Lizard.}
    \label{fig:sample-images}
\end{figure}

AnimalCLEF is part of the LifeCLEF lab, which aims to automate identification and understanding of life forms in the field of biodiversity informatics, implementing machine learning and computer vision methods \cite{lifeclef2026}.  
LifeCLEF is part of the Conference and Labs of the Evaluation Forum (CLEF), which is structured into two parts: evaluation labs like LifeCLEF and the peer-reviewed Conference \cite{CLEF2026}.
AnimalCLEF 2026 changed the task from the 2025 open-set assignment setting to a clustering setting across four species, including an unlabeled Texas horned lizard discovery set. 
In AnimalCLEF 2025, competitors were challenged with designing a model that would determine whether the depicted animal was new (not present in the training set) or known (where its identity must be provided) \cite{AnimalCLEF2025}. 
AnimalCLEF 2026 expanded the competition by focusing on individual-animal clustering across multiple species, including a discovery-only Texas horned lizard subset, and by evaluating submissions with the Adjusted Rand Index (ARI) \cite{animal-clef-2026,animalclef2026overview}. 
Our system placed 5th out of 230 teams in AnimalCLEF 2026 by combining global descriptors, local feature matching, learned pair scoring, and conservative graph construction across the four species. This approach demonstrates that calibrated graph construction, not just stronger descriptors, is central to reliable clustering-based re-identification.
\section{Related Work}

Animal re-identification aims to recognize unique animals across images by using traits such as coat patterns, spots, scale patterns, body texture, or other stable natural markings \cite{cermak2024wildlifedatasets,adam2025wildlifereid10k}. Animal re-identification enables estimates of population size, migration, and behavior by linking sightings of the same individual across time and sites. Field images vary in pose, lighting, background, and resolution. Prior AnimalCLEF work treats background, pose, lighting, and segmentation as robustness concerns because models can rely on capture conditions instead of identity markings when transferred to new images \cite{pakhomov2025hybrid,kim2025fusion}.

More recent AnimalCLEF approaches start with global image descriptors that convert each image into a feature vector and place visually similar animals close together. Domain-specific models such as MegaDescriptor are trained on animal re-identification datasets and have outperformed general-purpose vision models in individual-animal matching studies \cite{cermak2024wildlifedatasets,miyaguchi2025triplet,semenova2025meta}. The DS@GT 2025 pipeline compared DINOv2 and MegaDescriptor and found that triplet learning was more effective when applied to MegaDescriptor, a model already trained for animal re-identification \cite{oquab2024dinov2,miyaguchi2025triplet}. Other work has explored MiewID as a multi-species embedding model, where training across several species improved candidate retrieval for previously unseen animals \cite{otarashvili2024multispecies,conservationxlabs2024miewidmsv3,semenova2025meta}.

Because global descriptors summarize the entire image, they may miss small but important local identity cues. AnimalCLEF 2025 participants combined global retrieval with local feature matching. WildFusion calibrates and fuses global similarity scores, such as MegaDescriptor or DINOv2, with local matching scores from methods such as LoFTR and LightGlue \cite{cermak2024wildfusion,lindenberger2023lightglue, sun2021loftr}. The first-place hybrid global-local system first used global similarity to find likely matches, then used a weighted fusion of local matching methods to compare fine details such as markings and body patterns \cite{pakhomov2025hybrid}. Similarly, multilevel feature fusion methods combine deep image features with keypoint-based features, then use a threshold to decide whether an animal is known or unknown \cite{tan2025multilevel}. Fusion pipelines using MegaDescriptor, ALIKED, EVA02, WildFusion calibration, and test-time augmentation reported gains from combining global, local, and semantic features across species \cite{kim2025fusion}.

Prior systems also learned calibrated pairwise similarity scores. Gradient-boosting approaches combined pretrained embeddings, pairwise comparison features, and supervised decision boundaries to determine whether two images depicted the same animal \cite{fedorchenko2025gradient}. Other pipelines used WildFusion, XGBoost, and ArcFace-based fine-tuning to combine different types of information, including global descriptors, local matches, neighbor context, and species-specific models \cite{semenova2025meta}.

The 2026 clustering setting makes the task much more challenging than in the previous open-set assignment task. Prior systems relied on global retrieval, local verification, and learned pair scoring; the clustering setting also requires deciding which pair signals to treat as identity edges. A high-scoring false pair can be harmless if rejected, but damaging if it becomes a bridge edge inside a transitive graph.

\section{Data Overview}

The four animals featured in this year's competition have been selected to demonstrate the diversity of animal photography and the associated challenges of re-identification.
There are images on land and at sea, in various lighting conditions and perspectives, with and without labels for training.
Figure \ref{fig:sample-images} presents just a few of the images within the dataset to give a flavor of the unique traits that the animals possess.

\begin{figure}[htbp]
    \centering
    \includegraphics[width=\linewidth]{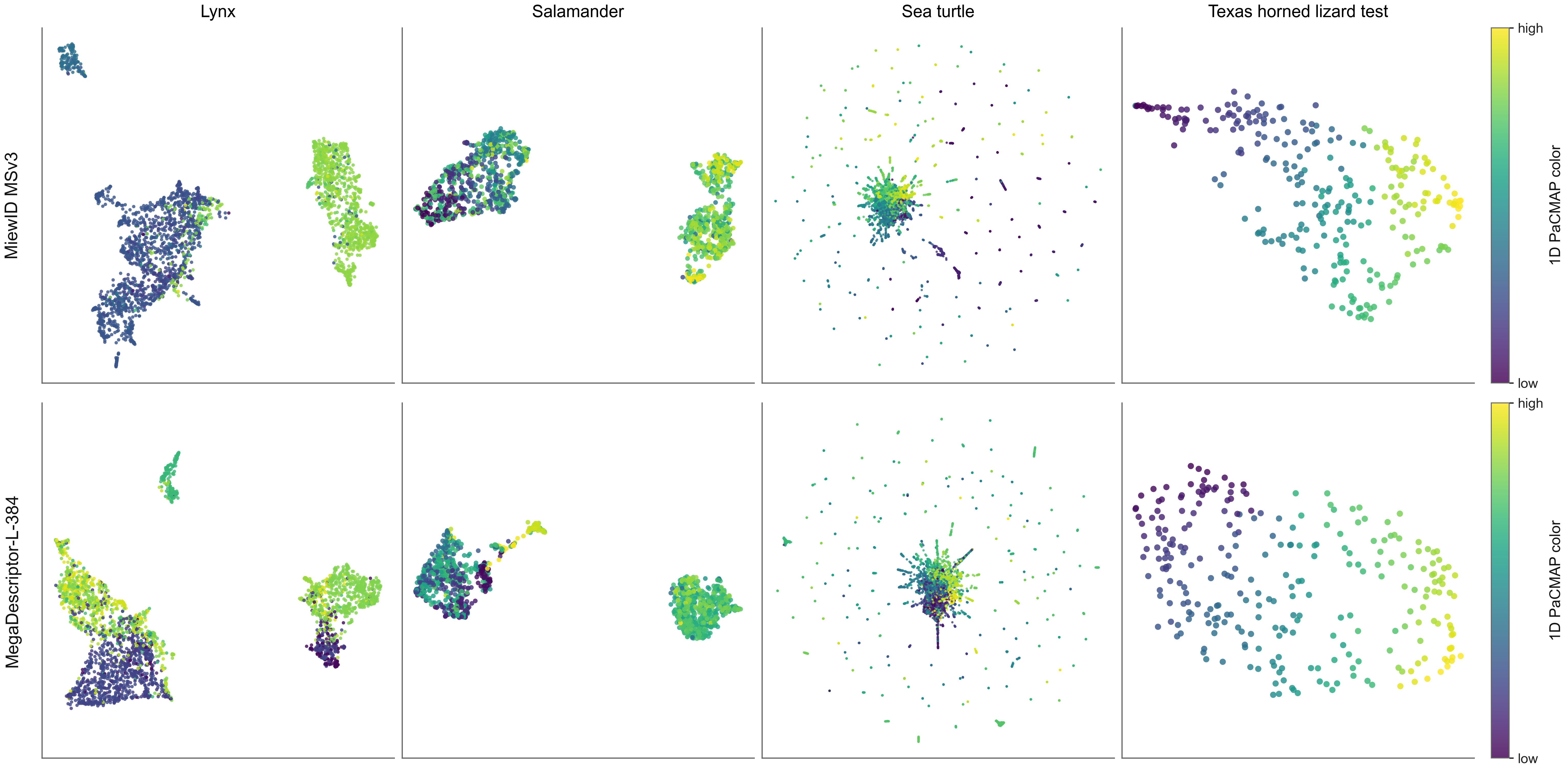}
    \caption{
        Visualization of clustering behavior for different animal species per MegaDescriptor and MiewID.
        The clustering behavior observed in Lynxes is likely due to a shift between day and night imaging.
        The clustering behavior of salamanders likely stems from images taken from a natural top-down perspective rather than those captured and held in hand.
        Sea turtles form a large, connected component near the center, with many surrounding clusters, likely due to the nature of thumbnail-sized images in the dataset.
        The distribution over horned lizards demonstrates a smooth topology in the localized embedding space which reflects the consistent photography framing.
    }
    \label{fig:pacmap-grid}
\end{figure}

The low-dimensional embedding structure in Figure~\ref{fig:pacmap-grid} mirrors these per-species traits and challenges. Lynxes are pre-segmented images from a camera-trap dataset and include stills extracted from videos taken in daylight, alongside low-light camera footage from nights.
The main identifier is the spotting and striping pattern on the fur, which is visible in many of the images.
The primary challenge is a significant domain shift in lighting, as many low-light images are difficult to differentiate.
Salamanders are photographed in a natural forest environment, with some photos of the backs showing unique yellow spots and glossy skin, and a few images taken with hands in the frame.
The turtle datasets are taken underwater, primarily capturing the head and scale-like patterns that uniquely identify individuals \cite{adam2024seaturtleid2022}.
This dataset contains many thumbnails and other low-quality images that are unlikely to be informative for re-identification.
The horned lizard's images are framed, showing the animal's belly scales held up by hand.
There are a few dozen clusters of darkened scales along the lizard's belly that appear to be its primary identifying markers.
The challenge for the lizards is that there are no labeled individuals in this set, and thus it is a pure clustering challenge.

\begin{table}[t]
  \centering
  \caption{Per-dataset statistics for the AnimalCLEF 2026 corpus.
  Identity statistics are computed on the training (database) split only.
  TexasHornedLizards has no training data (discovery-only task).}
  \label{tab:dataset-stats}
  \begin{tabular}{lrrrrrr}
  \toprule
  Dataset & Train & Test & IDs & Median & Max & Singletons \\
  \midrule
  LynxID2025         & 2{,}957 & 946 &  77 & 17 & 353 &  7.8\,\% \\
  SalamanderID2025   & 1{,}388 & 689 & 587 &  1 &  12 & 52.8\,\% \\
  SeaTurtleID2022    & 8{,}729 & 500 & 438 & 13 & 190 &  0.2\,\% \\
  TexasHornedLizards &     --- & 274 & --- & --- & --- &     --- \\
  \midrule
  Total              & 13{,}074 & 2{,}409 & 1{,}102 & --- & --- & --- \\
  \bottomrule
  \end{tabular}
\end{table}

Alongside these visual differences, the species also differ in their statistical structure (Table \ref{tab:dataset-stats}). Some species have fewer examples than others, and salamanders in particular contain many singleton identities with only one labeled training image. At roughly 13{,}000 training and 2{,}400 test images, the corpus is small enough to run on Kaggle notebooks or a personal machine. 
\section{Methodology}

\begin{figure}[!htbp]
    \centering
    \includegraphics[width=\linewidth]{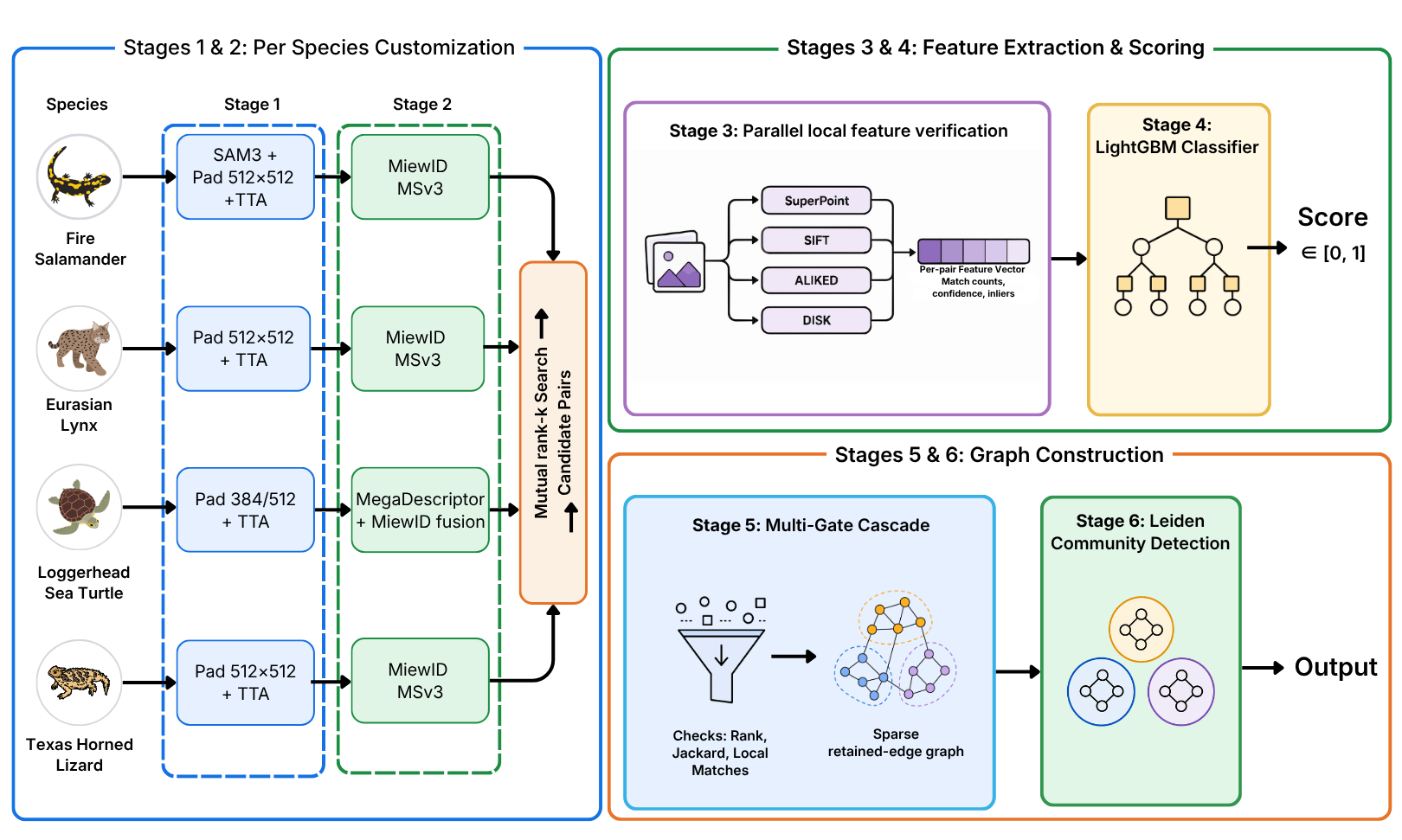}
\caption{
    \textbf{Species-aware submission pipeline.} 
    The system follows a fixed sequence: species-specific preprocessing and global representation, mutual-rank candidate retrieval, LightGlue local verification \cite{lindenberger2023lightglue}, LightGBM pair scoring \cite{ke2017lightgbm}, edge-admission gates, and graph clustering.}
    \label{fig:pipeline-overview}
\end{figure}

We model individual animal re-identification as species-specific graph construction over candidate image pairs. 
For each species, the system treats test images as nodes, proposes likely same-individual pairs, scores them using global, local, and neighborhood evidence, and converts retained edges into predicted identity clusters. 
Cross-species edges are not considered. The pipeline consists of species-specific preprocessing, candidate retrieval, pair scoring, edge admission, and graph clustering (Figure~\ref{fig:pipeline-overview}).

\subsection{Preprocessing and Global Representations}

Image preprocessing was selected separately for each species before retrieval and local verification. 
We used square padding, cropped black borders around Lynx images to keep the animal in view, applied vertical-flip test-time augmentation, and used SAM3 masks to isolate salamanders from noisy backgrounds \cite{carion2025sam3}.

Because the four species differ in pose, background structure, marking scale, and supervision, representation choices are evaluated by species rather than through a single universal backbone. 
The global backbones are MegaDescriptor-L-384 \cite{cermak2024wildlifedatasets} and MiewID-MSv3 \cite{otarashvili2024multispecies,conservationxlabs2024miewidmsv3}. 
We also tested the fusion of edge-level evidence from both backbones.

\subsection{Candidate Pair Retrieval and Neighborhood Context}

Candidate pair retrieval reduces the otherwise quadratic set of possible image pairs to a smaller shortlist. 
For each species, the nearest-neighbor search first keeps a neighborhood around each image. A second, usually smaller budget decides which candidate pairs receive expensive local verification and pair scoring. 
During training, a separate budget controls how many labeled neighbor pairs are mined to provide positive examples and hard negatives for the pair scorer. 
Test-time budgets instead determine which unlabeled pairs are eligible to become graph edges.

The retrieval stage also produces neighborhood-context features used downstream. 
For a candidate pair, these include retrieval rank, reciprocal-neighbor indicators, shared-neighbor counts, and Jaccard overlap between local neighborhoods. 
These features encode whether direct visual similarity agrees with the surrounding neighborhood structure.

\subsection{Local Verification and Pair Features}

Global descriptors summarize the full image, but individual identity often depends on local markings, spots, texture, or scale patterns. 
Shortlisted pairs receive local-verification features from LightGlue matching with SuperPoint, SIFT, ALIKED, and DISK keypoints \cite{lindenberger2023lightglue,detone2018superpoint,lowe2004distinctive,zhao2023aliked,tyszkiewicz2020disk}. 
Each branch compares the two images at a fixed working resolution and returns match-count and match-confidence summaries of the two images, including mean confidence, total support, confidence ratios, and thresholded support counts.

Local matching does not directly assign identities. 
Instead, it adds pair-level evidence about whether two shortlisted images share the same markings, spots, or texture. A pair can look similar under the global embedding model but still be rejected if local matching and neighborhood context are weak. 
Conversely, a lower-ranked retrieval pair can still be kept if local matching provides strong support and the surrounding graph evidence is consistent with a repeated individual.

\subsection{Learned Pair Scoring}

The pair scorer estimates whether a candidate image pair depicts the same individual. 
We train a LightGBM gradient-boosted decision-tree model \cite{ke2017lightgbm} on labeled candidate pairs from the training identities, using global similarity features, retrieval-rank features, neighborhood-context features, and local-verification summaries. 
Positive examples share a training identity, while negative examples come from different identities. 
Per-image caps and rank-band sampling are used to reduce class imbalance and expose the model to hard negatives rather than only visually unrelated pairs.

Pair scores are not final identity predictions. 
They are edge-strength proposals for graph construction. 
One false-positive edge can merge two otherwise separate identity components, whereas one false-negative edge may split a repeated individual into smaller clusters. 
The graph stage applies additional gates to convert pair scores into retained identity edges.

\subsection{Graph Construction and Clustering}

Graph construction converts scored candidate pairs into a sparse graph. 
Images are nodes, candidate same-individual relationships are proposed edges, and retained edges are selected using pair score, retrieval rank, local support, and neighborhood consistency. 
The edge filters include high-confidence core thresholds, lower-expansion thresholds, rank caps, local-support gates, common-neighbor and Jaccard gates, and component-shape controls, such as degree or component caps, when configured.

After edge filtering, the retained graph is clustered to produce predicted identity labels. 
Connected components treat every retained edge as a hard union. 
Leiden clustering can instead split a weighted graph into communities when edge strengths vary within a connected component \cite{traag2019leiden}. 
Backend comparisons are meaningful only when the edge set and weights being clustered are comparable.

\subsection{Final Competition Configuration}

The final competition system used species-specific operating points rather than a single global setting. This reflected the structure of the four datasets: LynxID2025 required conservative graph growth across camera-trap domains, SalamanderID2025 contained many singleton identities where false bridge edges were costly, SeaTurtleID2022 favored stable global retrieval under low-resolution imagery, and TexasHornedLizards lacked labeled identities for supervised pair mining. Across species with local verification, we summarized four LightGlue branches, SuperPoint, SIFT, ALIKED, and DISK, using match-count and confidence features in a LightGBM pair scorer. The scorer used 250 trees, learning rate 0.06, 63 leaves, row and column subsampling of 0.8, and $\ell_2$ regularization of 0.5; its output was treated as graph-edge evidence rather than as a direct identity classifier.

\begin{table*}[!htbp]
\centering
\caption{\textbf{Species-specific settings in the selected competition-period system.} Pair budgets are shown as retrieved/scored candidate counts.}
\label{tab:final-species-config}
\scriptsize
\setlength{\tabcolsep}{3pt}
\begin{tabular}{@{}>{\raggedright\arraybackslash}p{0.17\linewidth}>{\raggedright\arraybackslash}p{0.22\linewidth}>{\raggedright\arraybackslash}p{0.27\linewidth}>{\raggedright\arraybackslash}p{0.25\linewidth}@{}}
\toprule
Species & Representation and preparation & Candidate and training budgets & Graph setting and final role \\
\midrule
LynxID2025 & MiewID, 512 px; vertical flip & test 40/20; train 30/20, top-48 & core/expand 0.93/0.88; Leiden 0.02, edge floor 0.82; at least two shared neighbors; known-identity attachment threshold 0.94, margin 0.04, support 2; retained baseline \\
SalamanderID2025 & MiewID, 512 px; SAM3 foreground mask and square pad; vertical flip & test 224/128 with 50 context neighbors; train 80/40, top-48; graph 60/30 with 60 context neighbors and top-72 tuning & core/expand/very-core 0.9618/0.9255/0.9811; singleton/merge 0.9829/0.9546; Leiden 0.03, edge floor 0.80, seed 7, eight iterations; promoted replacement \\
SeaTurtleID2022 & MegaDescriptor, 384 px; vertical flip & test 50/25; train 30/30, top-48 & core/expand 0.92/0.85; Leiden 0.02, edge floor 0.81; at least two shared neighbors; retained baseline \\
TexasHornedLizards & MiewID, 512 px & test 60/30; no supervised train-pair mining & core/expand 0.95/0.90; at least three shared neighbors; known-identity attachment disabled; retained baseline \\
\bottomrule
\end{tabular}
\end{table*}

Table~\ref{tab:final-species-config} summarizes the selected per-species operating points. The selected full system retained the baseline configurations for LynxID2025, SeaTurtleID2022, and TexasHornedLizards, and replaced only the SalamanderID2025 predictions with the more selective SAM3, vertical-flip, MiewID, and Leiden safe-graph configuration. This was the main empirical lesson from our ablations: improvements did not transfer uniformly across species, so the final configuration selected the species-specific operating point with the strongest full-system behavior rather than promoting every exploratory improvement. The code can be found at \url{https://github.com/dsgt-arc/animalclef-2026}.

\subsection{Evaluation Protocol}

Public and private leaderboard scores are aggregate ARI values for complete submissions. The public score is computed on the public portion of the hidden test set, while the private score is computed on the remaining hidden test data; the private leaderboard uses approximately 52\% of the test images. 
The Adjusted Rand Index used for scoring is defined in Eq.~\ref{eq:ari}.

Local validation used identity-grouped cross-validation on the labeled training images. Training identities, not individual images, were assigned to folds, so images of the same known animal could not appear in both the training and validation portions of a fold. For each validation fold, the system predicted clusters for the held-out images and compared them with the known identity labels using ARI. 
These local CV scores were used to compare candidate settings before leaderboard submission.

Ablation and sensitivity studies differ in what they hold fixed. Some studies change one component, while others remove a component without retraining the pair scorer or retuning graph thresholds. 
We therefore interpret these studies as sensitivity evidence unless the table or text states that the scorer was retrained and graph settings were retuned.

\section{Results}

\subsection{Main Competition Results}
AnimalCLEF 2026 evaluates submissions with the Adjusted Rand Index (ARI), a chance-corrected agreement measure between the hidden identity partition $U$ and the predicted cluster partition $V$ \cite{animal-clef-2026,animalclef2026overview,hubert1985comparing}. Let $n_{ij}=|U_i \cap V_j|$, $a_i=\sum_j n_{ij}$, $b_j=\sum_i n_{ij}$, $n=\sum_{ij} n_{ij}$, and $\binom{x}{2}=x(x-1)/2$. The ARI is

\begin{equation}
\mathrm{ARI} =
\frac{
\sum_{ij}\binom{n_{ij}}{2}
-
\frac{\sum_i \binom{a_i}{2}\sum_j \binom{b_j}{2}}{\binom{n}{2}}
}{
\frac{1}{2}\left[
\sum_i \binom{a_i}{2}
+
\sum_j \binom{b_j}{2}
\right]
-
\frac{\sum_i \binom{a_i}{2}\sum_j \binom{b_j}{2}}{\binom{n}{2}}
}.
\label{eq:ari}
\end{equation}

A score of 1 indicates identical partitions, values near 0 indicate chance-level agreement, and negative values indicate worse-than-chance agreement. In this challenge, false splits and false merges both reduce ARI.

The strongest competition-period DS@GT result scored 0.73275 public and 0.67424 private ARI, placing 5th on the final private leaderboard among 230 teams \cite{animal-clef-2026-leaderboard}. The LynxID2025, SeaTurtleID2022, and TexasHornedLizards blocks were retained from the frozen baseline, while SalamanderID2025 was the only promoted replacement, using MiewID retrieval, SAM3 mask-square-pad preprocessing, vertical-flip augmentation, local verification, raw LightGBM pair scoring, and evidence-gated Leiden clustering. Later, SeaTurtle and Lynx variants showed that stronger local or post-deadline operating points could trade off public and private performance, but they did not yield a better official-period all-species submission. All reported public/private values are full-submission ARI, not hidden per-species ARI.

Table~\ref{tab:leaderboard} anchors this result against the rest of the competition. The five highest-ranked teams on the private leaderboard fell within a narrow band of about $0.030$ ARI ($0.674$--$0.704$), with DS@GT trailing fourth place by roughly $0.001$ ARI, while all five scored approximately three times higher than the organizer WildFusion and MegaDescriptor baselines ($0.206$--$0.212$)~\cite{animal-clef-2026-leaderboard}. The challenge was therefore both difficult, with baselines clustered near $0.21$, and closely contested among the leading systems.

\begin{table}[htbp]
  \centering
  \caption{\textbf{AnimalCLEF 2026 final (private) leaderboard: top five teams and organizer baselines.} The private leaderboard determines the official ranking. DS@GT (team \emph{DS@GT LifeCLEF}) placed 5th of 230 teams on the private split and 3rd on the public split (public ARI $0.73275$). Organizer baseline entries are listed for reference and are not ranked. Source: Kaggle leaderboard~\cite{animal-clef-2026-leaderboard}.}
  \label{tab:leaderboard}
  \small
  \setlength{\tabcolsep}{6pt}
  \begin{tabular}{@{}clrr@{}}
  \toprule
  Rank & Team & Private ARI & Entries \\
  \midrule
  1 & M.I.A & 0.70393 & 270 \\
  2 & Ram Lab & 0.70050 & 154 \\
  3 & VIPL-VSU & 0.69632 & 88 \\
  4 & HSU Lab & 0.67537 & 174 \\
  \textbf{5} & \textbf{DS@GT LifeCLEF (ours)} & \textbf{0.67424} & \textbf{37} \\
  \midrule
  \multicolumn{4}{@{}l}{\emph{Organizer baselines}} \\
  --- & WildFusion + HDBSCAN & 0.21221 & 1 \\
  --- & MegaDescriptor + MiewID + DBSCAN & 0.20646 & 1 \\
  \bottomrule
  \end{tabular}
\end{table}

\begin{figure*}[!htbp]
    \centering
    \includegraphics[width=\linewidth]{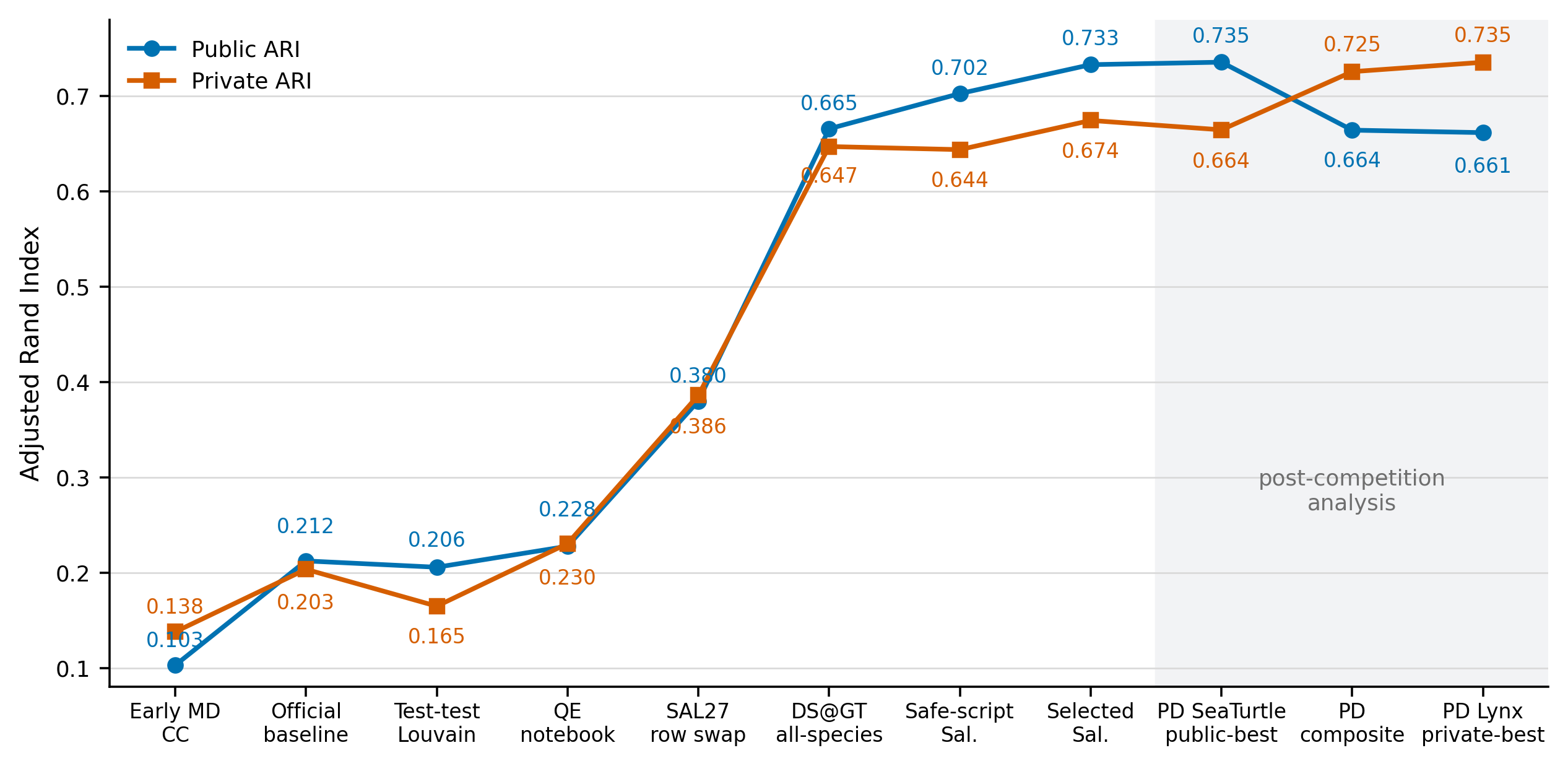}
    \caption{\textbf{Public/private submission score trajectory.} Scores include early MegaDescriptor and test-to-test graph submissions, the DS@GT all-species system, the selected Salamander replacement, and post-deadline analyses. All scores are full-submission ARI. Shaded points indicate submissions scored after the official competition closed.}
    \label{fig:score-ladder}
\end{figure*}

Public/private scores progressed from early MegaDescriptor without trained pair scorers to the DS@GT all-species system and the selected Salamander replacement, which produced the highest competition-period public score (Figure~\ref{fig:score-ladder}). Post-deadline SeaTurtle improved public score without improving private score, while wider Lynx graphs improved private score at public-score cost.

\FloatBarrier

\subsection{Global Retrieval and Backbone Selection}

\begin{figure*}[!htbp]
    \centering
    \includegraphics[width=\textwidth]{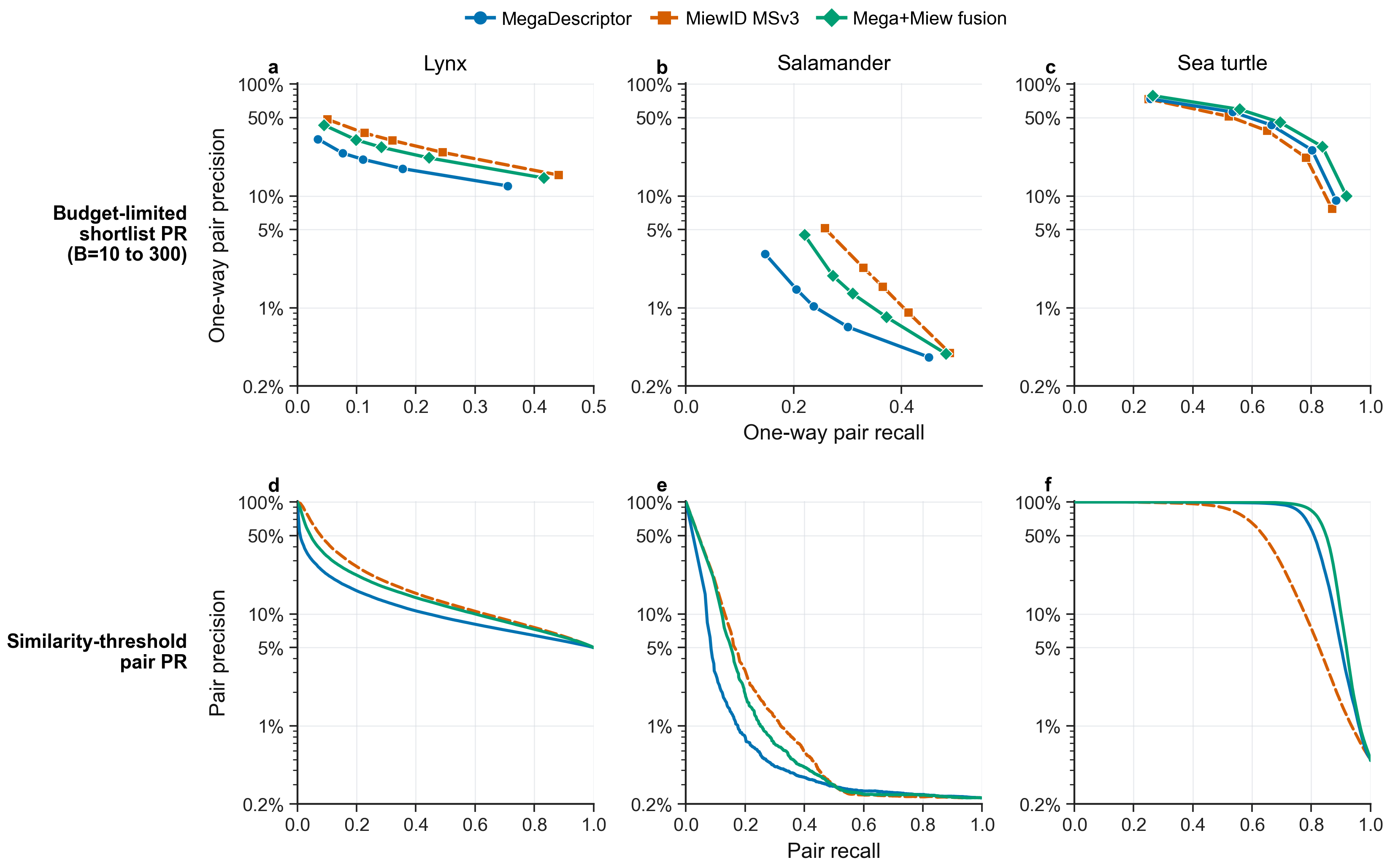}
    \caption{\textbf{Global backbone precision-recall tradeoffs by species.} Top panels show fixed-budget shortlist retrieval; bottom panels show threshold-sweep retrieval before local verification or graph filtering.}
    \label{fig:global-pr-matrix}
\end{figure*}

Retrieval metrics differed by species: MiewID MSv3 \cite{conservationxlabs2024miewidmsv3,otarashvili2024multispecies} produced stronger curves for Lynx and Salamander than MegaDescriptor \cite{cermak2024wildlifedatasets} in the evaluated budgets, while SeaTurtle retrieval was strongest when MegaDescriptor and MiewID evidence were fused (Figure~\ref{fig:global-pr-matrix}; Tables~\ref{tab:global-budget-pr}--\ref{tab:global-score-threshold-pr}). The final ARI still relied on local verification, edge admission, and clustering.

\begin{table*}[!htbp]
\centering
\caption{\textbf{Fixed-budget global shortlist precision-recall by species.} Cells give one-way pair recall/precision as percentages rounded to one decimal. Bold model names mark the strongest displayed retrieval model for each species across the shown budgets.}
\label{tab:global-budget-pr}
\small
\setlength{\tabcolsep}{2.8pt}
\begin{tabular}{@{}llccccc@{}}
\toprule
Species & Model & \multicolumn{5}{c}{Retrieval budget $B$; recall / precision (\%)} \\
\cmidrule(l){3-7}
& & 10 & 30 & 50 & 100 & 300 \\
\midrule
\multirow{3}{*}{Lynx} & MegaDescriptor & 3.5\% / 32.3\% & 7.7\% / 24.1\% & 11.1\% / 21.2\% & 17.8\% / 17.5\% & 35.5\% / 12.3\% \\
& \textbf{MiewID MSv3} & 5.1\% / 48.3\% & 11.3\% / 36.5\% & 16.0\% / 31.3\% & 24.5\% / 24.5\% & 44.1\% / 15.4\% \\
& Mega+Miew fusion & 4.5\% / 43.0\% & 10.0\% / 31.7\% & 14.2\% / 27.2\% & 22.2\% / 21.9\% & 41.6\% / 14.5\% \\
\addlinespace[2pt]
\multirow{3}{*}{Salamander} & MegaDescriptor & 14.7\% / 3.0\% & 20.5\% / 1.5\% & 23.7\% / 1.0\% & 30.1\% / 0.7\% & 45.1\% / 0.4\% \\
& \textbf{MiewID MSv3} & 25.8\% / 5.1\% & 33.0\% / 2.3\% & 36.6\% / 1.6\% & 41.3\% / 0.9\% & 49.0\% / 0.4\% \\
& Mega+Miew fusion & 22.0\% / 4.5\% & 27.3\% / 1.9\% & 31.0\% / 1.3\% & 37.2\% / 0.8\% & 48.3\% / 0.4\% \\
\addlinespace[2pt]
\multirow{3}{*}{SeaTurtle} & MegaDescriptor & 25.6\% / 73.7\% & 53.4\% / 56.4\% & 66.5\% / 43.0\% & 80.2\% / 25.7\% & 88.4\% / 9.2\% \\
& MiewID MSv3 & 25.0\% / 73.5\% & 52.2\% / 51.4\% & 65.2\% / 38.2\% & 78.3\% / 22.0\% & 87.0\% / 7.7\% \\
& \textbf{Mega+Miew fusion} & 26.5\% / 78.7\% & 55.8\% / 59.4\% & 69.5\% / 45.4\% & 83.7\% / 27.6\% & 91.9\% / 9.9\% \\
\bottomrule
\end{tabular}
\end{table*}

\begin{table*}[!htbp]
\centering
\caption{\textbf{Score-threshold global pair-retrieval summary.} Pair prior and candidate counts are species-level. AP values are rounded to three decimals; bold marks the strongest threshold curve within each species.}
\label{tab:global-score-threshold-pr}
\small
\setlength{\tabcolsep}{5pt}
\begin{tabular}{@{}lccccc@{}}
\toprule
Species & Pair prior & Positive / candidate pairs & \multicolumn{3}{c}{Average precision} \\
\cmidrule(l){4-6}
& & & MegaDescriptor & MiewID MSv3 & Fusion \\
\midrule
Lynx & 0.050 & 218,226 / 4,370,446 & 0.123 & \textbf{0.198} & 0.169 \\
Salamander & 0.002 & 2,183 / 962,578 & 0.055 & \textbf{0.082} & 0.080 \\
SeaTurtle & 0.005 & 186,777 / 38,093,356 & 0.804 & 0.636 & \textbf{0.844} \\
\bottomrule
\end{tabular}
\end{table*}

\FloatBarrier

\subsection{Component Evidence}

Table~\ref{tab:feature-family-attribution} shows which kinds of evidence the LightGBM pair scorer relied on most when judging candidate image pairs \cite{ke2017lightgbm,lundberg2017shap}. Across species, retrieval-rank features carried the highest per-feature attribution by a wide margin, followed by the DISK LightGlue branch. The SuperPoint branch was most prominent for Salamander, while for SeaTurtle the Global cosine features (driven primarily by MiewID similarity) rose to the second strongest per-feature family. Neighborhood-context and SIFT features contributed the least per feature, consistent with a scorer that combines several signals rather than ranking pairs by a single embedding score.

\begin{table*}[!htbp]
\centering
\caption{\textbf{Feature-family attribution in the LightGBM pair scorer.} Each cell shows the average attribution of a single feature in that family when the scorer judges a candidate same-individual image pair, measured by mean$|$SHAP$|$ \cite{lundberg2017shap} and divided by the family's feature count so that wide families (the four LightGlue branches, WildFusion summaries) are directly comparable with narrow ones (Global cosine, Neighborhood context). Values are in the model's raw-output (log-odds) units, so a value of $0.10$ means each feature in that family on average shifts the pair-decision logit by about $\pm 0.10$ per pair --- larger is more influential. \textit{Mean} averages the three species. Bold marks the most-attributed family within each species column; rows are ordered by \textit{Mean}.}
\label{tab:feature-family-attribution}
\small
\setlength{\tabcolsep}{8pt}
\begin{tabular}{@{}lrrrr@{}}
\toprule
Feature family & Salamander & SeaTurtle & Lynx & Mean \\
\midrule
Retrieval rank       & \textbf{0.222} & \textbf{0.277} & \textbf{0.259} & 0.253 \\
LightGlue DISK       & 0.160          & 0.124          & 0.160          & 0.148 \\
Global cosine        & 0.097          & 0.143          & 0.074          & 0.105 \\
LightGlue SuperPoint & 0.140          & 0.085          & 0.085          & 0.103 \\
\addlinespace[0.25em]
WildFusion summaries & 0.062          & 0.067          & 0.056          & 0.062 \\
LightGlue ALIKED     & 0.053          & 0.046          & 0.048          & 0.049 \\
Neighborhood context & 0.043          & 0.034          & 0.041          & 0.039 \\
LightGlue SIFT       & 0.038          & 0.033          & 0.041          & 0.037 \\
\bottomrule
\end{tabular}
\end{table*}

\FloatBarrier

\begin{table*}[!htbp]
\centering
\caption{\textbf{Ablations grouped by species.} Public/private scores are full-submission ARI, rounded to three decimals. Each species block is headed by its selected submission (reference); subsequent rows change or remove a single component.}
\label{tab:component-removal}
\small
\setlength{\tabcolsep}{5.5pt}
\begin{tabular}{@{}>{\raggedright\arraybackslash}p{0.50\linewidth}rr@{}}
\toprule
Reference or change & Public & Private \\
\midrule
\multicolumn{3}{@{}l}{\textbf{SalamanderID2025}} \\
\addlinespace[0.15em]
Best submission (reference) & \textbf{0.733} & \textbf{0.674} \\
Remove SAM3 mask-square-pad & 0.660 & 0.629 \\
MegaDescriptor instead of MiewID & 0.650 & 0.637 \\
Remove vertical-flip TTA & 0.714 & 0.663 \\
\midrule
\multicolumn{3}{@{}l}{\textbf{SeaTurtleID2022}} \\
\addlinespace[0.15em]
Best submission (reference) & \textbf{0.688} & \textbf{0.630} \\
MiewID only & 0.684 & 0.629 \\
MegaDescriptor+MiewID edge mean & 0.679 & 0.616 \\
Remove vertical-flip TTA & 0.670 & 0.618 \\
\midrule
\multicolumn{3}{@{}l}{\textbf{LynxID2025}} \\
\addlinespace[0.15em]
Best submission (reference) & \textbf{0.702} & \textbf{0.703} \\
Remove vertical-flip TTA & 0.656 & 0.616 \\
\bottomrule
\end{tabular}
\end{table*}
\FloatBarrier

Salamander was most sensitive to the global representation and foreground pre-processing: replacing MiewID with MegaDescriptor and removing the SAM3 mask-square-pad pre-processing both reduced the public and private scores, while removing vertical-flip augmentation caused a smaller but consistent loss. SeaTurtle showed the opposite pattern: although retrieval evidence favored MegaDescriptor+MiewID fusion, the downstream replacement controls favored MegaDescriptor alone, suggesting that stronger retrieval did not automatically translate into better graph clustering. Removing TTA from SeaTurtle lowered the public score with little change to the private score. For Lynx, removing vertical-flip TTA reduced the reference from $0.702/0.703$ to $0.656/0.616$. Across species, the LightGlue match families also carried substantial attribution in the learned pair scorer (Table~\ref{tab:feature-family-attribution}), consistent with local evidence being a useful graph-construction signal.

\subsection{Edge Admission and Cluster Shape}

\begin{figure*}[!htbp]
    \centering
    \includegraphics[width=\linewidth]{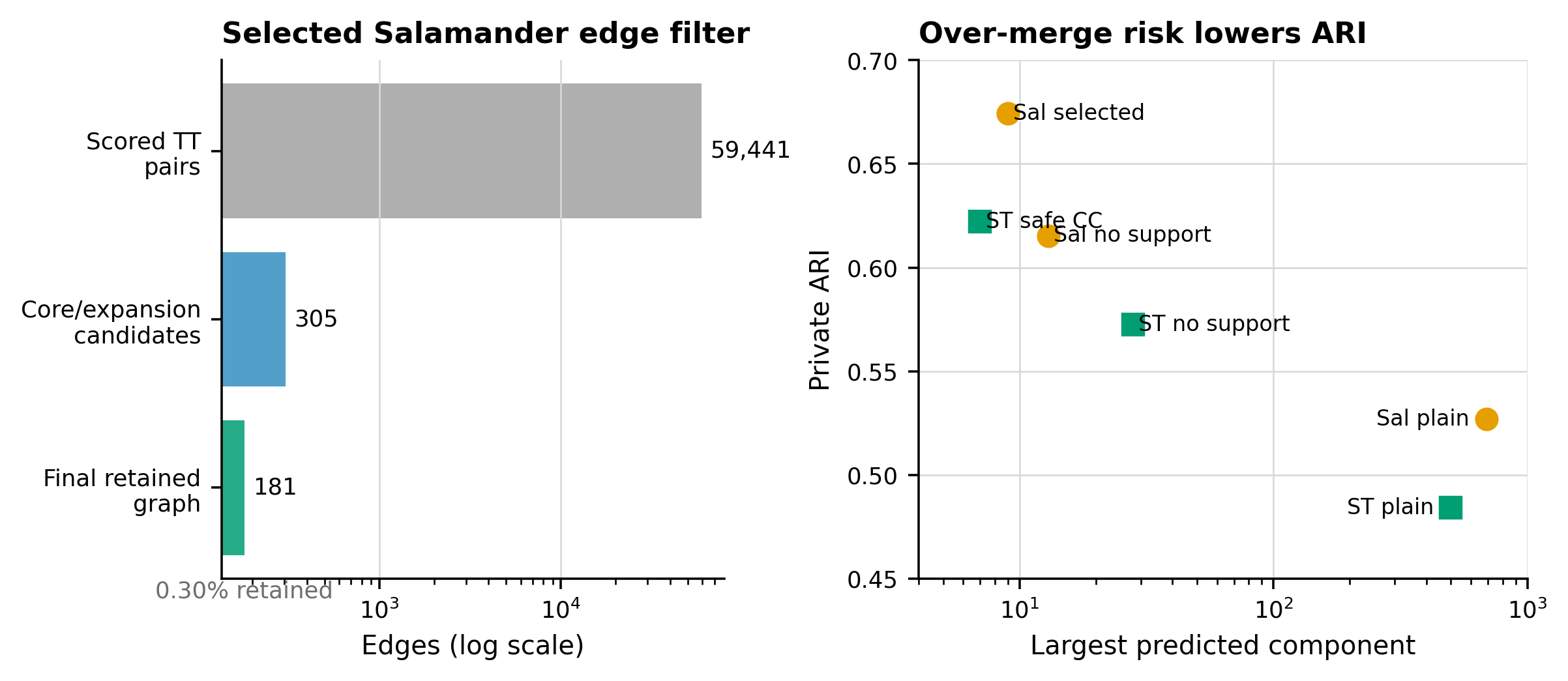}
    \caption{\textbf{Edge admission and over-merge risk.}
    The selected Salamander edge-admission filter retains a small subset of the candidate-pair universe, while ungated threshold graphs and local-support gate removals produce larger components and lower private ARI. The largest failures occur when threshold graphs collapse many images into one dominant component.}
    \label{fig:graph-safety}
\end{figure*}

Graph construction changed ARI by controlling retained edge sets and component sizes. Plain threshold graphs and support-gate removals produced graph shapes consistent with bridge-edge over-merging, creating large transitive components and reducing ARI. A post-deadline backend-isolation check that ran connected components and Leiden on matched retained edge sets found lower transitive over-merging with Leiden, most clearly for SeaTurtle, but leaderboard scores remained mixed. The selected Salamander edge filter retained a small graph, whereas ungated threshold and support-gate removal variants produced larger components associated with lower private ARI (Figure~\ref{fig:graph-safety}).

The selected Salamander submission reduced fragmentation relative to the frozen baseline while keeping the largest component small: clusters fell from 599 to 574, singletons from 549 to 523, and same-cluster pairs rose from 158 to 260. The private-strong widened-CC Lynx analysis instead used a 311-image largest component and 55,447 same-cluster pairs, coinciding with higher private ARI and lower public ARI.

\section{Discussion}

Animal re-identification in AnimalCLEF 2026 depended on edge admission as much as descriptor quality. Stronger retrieval increased both true same-individual candidates and plausible false candidates; once accepted as graph edges, false pairs could propagate through transitive closure or community detection. The selected submission combined broad candidate generation with conservative graph construction, producing more non-singleton structure while keeping the largest predicted component small (Figure~\ref{fig:graph-safety}). K-reciprocal reranking \cite{zhong2017reranking} was also tested as a post-retrieval refinement, but did not improve ARI for any species.

The strongest competition-period result came from the selected Salamander submission. It scored 0.733 public / 0.674 private ARI and placed DS@GT 5th out of 230 teams. Relative to the frozen baseline, salamander reduced fragmentation while preserving a conservative graph shape: clusters fell from 599 to 574, singletons fell from 549 to 523, and the largest predicted component increased only from 7 to 9. The gain, therefore, came from adding supported same-individual links, not from letting a few large components absorb uncertain images.

Salamander also had an exploratory spot-based branch. We tested HSV color thresholding, SAM3 prompts for yellow spots, spot masks, centroid geometry, constellation and Delaunay features, and Kornia-based keypoint matching. These probes were motivated by the species' distinctive markings, but they remained exploratory. Spot-level cues may be useful, but only when they improve calibrated pair evidence and edge admission.

SeaTurtle retrieval metrics did not translate reliably into final clustering performance. The fused MegaDescriptor and MiewID retrieval curve was strongest for candidate generation, but downstream graph controls did not consistently convert that advantage into private-score gains. Local support still mattered: disabling the local-support gate reduced the SeaTurtle graph-support candidate from 0.622 to 0.572 private ARI.. The final graph operating point, rather than retrieval alone, shaped transfer.

Lynx had the strongest public/private graph-shape divergence. In post-deadline Lynx analyses, using aggregate public/private scores, larger connected structures improved private ARI and reduced public ARI. Our Lynx graph-support replays were mostly neutral around the low accepted-edge connected-components reference, so no single local branch was globally preferable. The main Lynx uncertainty was the graph operating point: the public and private splits rewarded different component-size and edge-admission tradeoffs.

\FloatBarrier

\subsection{Lynx 3D}

One exploratory direction used pose estimators to re-project images and reduce the effects of occlusion and soft-body deformation. The hypothesis was that orientation information is implicitly captured in images, and that projecting animals into a canonical pose could help local feature matchers align individual-specific patterns in pixel space.

The first direction followed pelage-unwrapping \cite{algasov2025pelage}, in which fur texture is flattened onto a surface. The approach used monocular depth and surface-normal estimates, then optimized a projection that made the normals more consistent. Hyperbolic and cylindrical projections are both reasonable spaces for this goal because they can impose geometric consistency between pixels that appear on a curved animal body. We tried to reproduce this direction and used Lynx as the main testbed because Lynx have fur and are often less occluded than the other species. In practice, the resulting unwraps were not usable for downstream matching. The distortion was severe, and each image produced its own UV space rather than a shared space across images. The underlying model also relied on foreground-background contrast, whereas the Lynx images in the competition were already pre-segmented and many were captured at night, creating a likely domain shift.

\begin{figure*}[!htbp]
    \centering
    \includegraphics[width=\linewidth]{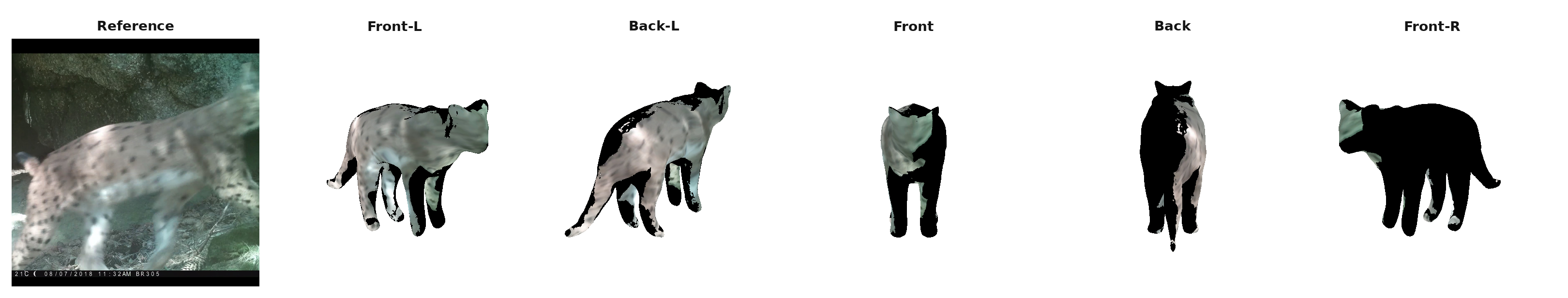}
    \caption{\textbf{Lynx 3D and UV-projection exploration.} 3D-Fauna-style reconstruction produced visually coherent single-image projections in some cases, but multi-image synthesis and downstream matching were limited by pose error, sparse surface coverage, and domain shift.}
    \label{fig:lynx3d-discussion}
\end{figure*}

The second direction used 3D-Fauna, a feed-forward model that reconstructs posed and textured 3D meshes from quadruped images \cite{li2024learning3dfauna}. Given a stack of Lynx image embeddings, we averaged the embeddings to obtain a single shared UV space. The intended benefit was that pixels from the original image could be projected onto a shared surface where the same identifying markings would occupy similar locations across pose and viewpoint changes.

In practice, pose and orientation were often plausible but difficult to validate quantitatively. The generated latent texture was not useful for re-identification because it mostly contained blurry color regions rather than high-frequency individual markings. Identifying information is likely nonlinear and high frequency, and it is also confounded by camera type, illumination, and viewpoint. Multi-image UV synthesis was also difficult: images of the same individual often came from different photographic distributions, and pose recovery errors accumulated when pixels from multiple images were projected into the same space. Gaussian point-cloud projection and related rendering attempts produced either high-frequency speckling from sparse surface coverage or blurry texture from uncertain correspondence. A single-image UV render was more visually coherent (Figure~\ref{fig:lynx3d-discussion}), but the resulting image was outside the training distribution of global embedding models.

We therefore treated UV projections as possible local-matching inputs rather than as replacement global embeddings. Local matchers can be more tolerant of unusual image geometry when the feature extractor captures repeatable visual structure; SIFT, for example, is not learned from the same wildlife-image distribution as the global embedding models. For Lynx, however, adding 3D-Fauna UV projections degraded overall model performance. The result is inconclusive rather than negative in a broad sense: better UV synthesis, better uncertainty handling, or a local-matcher analysis focused on spot-pattern stability could still make canonicalized texture useful.

\FloatBarrier

\subsection{SeaTurtle Upscaling}

SeaTurtle also received an exploratory 3D treatment, but a quadruped model is poorly matched to turtle heads. The goal was to model the head, where large-scale shaped markings and beak geometry carry much of the identity evidence. A separate shape-embedding parameterization had limited success. Even with a general-purpose feature extractor such as DINOv3 \cite{simeoni2025dinov3}, the model often failed to recover a useful orientation or pose when the head was represented as a semi-ellipsoid or related geometric surface. Background water was also projected onto the surface in unnatural ways, so this SeaTurtle 3D direction was not used in the final pipeline.

The 3D exploration still exposed useful preprocessing issues. We needed a segmentation model that could separate the turtle head from the background water. 
SAM3 center-object segmentation worked well for many images, but small thumbnail-sized images and full-body turtle poses led to degenerate cases.
We also tried upscaling very small images, often less than $50 \times 50$ pixels, because simple interpolation produced blurry results. 
Neural upscaling and JPEG artifact removal were considered as ways to recover line and scale-boundary evidence that might help turtle re-identification, but these steps were not promoted into the final system.

\subsection{Texas Horned Lizard Detections}

\begin{figure}[!htbp]
    \centering
    \includegraphics[width=0.4\linewidth]{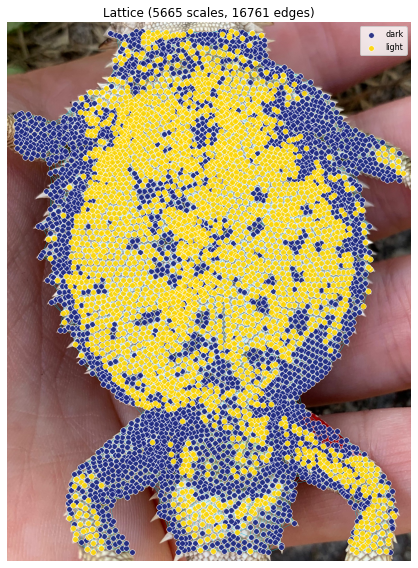}
    \caption{\textbf{Texas horned lizard scale-detection exploration.} Scale-level segmentation and Delaunay-style geometry could provide species-specific cues, but the current evidence is exploratory until connected to identity-clustering validation.}
    \label{fig:texas-scale-detector-discussion}
\end{figure}

Texas horned lizards posed challenges due to the lack of labeled training identities, which prevented the use of the same validation loop as for Lynx, Salamander, and SeaTurtle. A further direction, explored but not integrated into the final pipeline, detected individual scales in each horned lizard image.
The lizards were photographed with consistent framing and pose, and individual identification relied on their unique belly scaling patterns. This pattern forms a regular lattice, allowing for quantification of total area in units of scales and mapping of binary coloring on a shared template.
Several parametric methods, including edge sharpening and flood-fill techniques, were explored for scale extraction; however, variability in geometry and image consistency complicated this process. 
The labeling strategy involved normalizing image sizes to ensure a consistent pixel-to-unit-distance ratio, followed by cropping each image to a square grid ten units wide and manually labeling the visible scales.
A U-Net model was employed to predict scale locations, with its output used to pre-annotate subsequent annotation rounds.
This semi-supervised process produced a model capable of identifying individual scales (Figure~\ref{fig:texas-scale-detector-discussion}).

Although this model was not deployed in production, future work could leverage it to determine the diameter of the belly section and the relative positions of dark spots from the lizard's center. 
Subsequent methods could include point matching and custom fingerprinting.
The first method would use the detected points as features for algorithms such as LightGlue, aiming to find a bipartite matching between two point sets that preserves spatial locality between images.
The second approach would generate features from the images, such as U-Net embeddings, and directly cluster them, assuming that scale detection naturally clusters individuals.
This process may be further refined by generating a regular grid, normalizing dot locations, and comparing direct matches within this simplified geometry to enhance interpretability.

\FloatBarrier

\section{Future Work}
The highest-priority future work is controlled operating-point validation. Small changes in retrieval budget, edge-admission thresholds, and clustering backend changed whether candidate pairs became useful identity links or false bridge edges, so future studies should sweep threshold values under fixed retrieval caches and report ARI together with graph-shape diagnostics such as cluster count, singleton count, largest component, and same-cluster pairs. This validation protocol is especially important for SeaTurtle and Lynx, where retrieval quality and final clustering performance were not always aligned.

Once the operating point is controlled, two additions are natural to test. Texas horned lizards should be evaluated with a species-specific ventral-spot branch when usable ventral crops are available; Biffi et al.\ \cite{biffi2025texas} report that HotSpotter \cite{crall2013hotspotter} matching on ventral spot crops reaches 94\% match accuracy against PIT-tag and multilocus-genotype ground truth, with near-100\% accuracy when top-10 candidate review is allowed. A DINOv3 backbone fine-tuned on WildlifeReID-10k \cite{simeoni2025dinov3,adam2025wildlifereid10k} could also refresh the global descriptor, but it should be judged through the full retrieval-to-graph pipeline rather than retrieval metrics alone. Longer term, a Generalized Category Discovery-inspired formulation \cite{wen2023simgcd,zhang2025prompt} is worth exploring as an alternative to pair scoring followed by graph construction.

\section{Conclusions}

The DS@GT AnimalCLEF 2026 system treated clustering‑based animal re-identification as a species-aware graph construction problem rather than a single-descriptor ranking problem. The selected submission achieved 0.733 public ARI and 0.674 private ARI, placing 5th out of 230 teams. Post-deadline analyses linked SeaTurtle and Lynx transfer differences to local support, graph shape, and sensitivity to public/private split. Edge admission determined whether retrieved pairs formed a predicted identity structure or incurred over-merge risk.

\section*{Acknowledgements}

We thank the Data Science at Georgia Tech (DS@GT) CLEF competition group for their support.
This research was supported in part by research cyberinfrastructure resources and services provided by the Partnership for an Advanced Computing Environment (PACE) at the Georgia Institute of Technology, Atlanta, Georgia, USA \cite{PACE}. 

\section*{Declaration on Generative AI}
  
During the preparation of this manuscript, the authors used Grammarly to assist with grammar and spelling checks and ChatGPT to help improve clarity and revise sentence wording. The authors reviewed and edited all AI-assisted text as appropriate and take full responsibility for the final content of the publication.

\bibliography{main}
\end{document}